\documentclass[sigconf]{acmart}

\renewcommand\footnotetextcopyrightpermission[1]{} 
\AtBeginDocument{%
  }

\setcopyright{acmcopyright}
\copyrightyear{2018}
\acmYear{2018}
\acmDOI{XXXXXXX.XXXXXXX}

\acmConference[MOBILESoft 2024]{11th International Conference on Mobile Software Engineering and Systems 2024}{April 2024}{Lisbon, Portugal}
\acmPrice{15.00}
\acmISBN{978-1-4503-XXXX-X/18/06}

\usepackage{setspace}
\usepackage{algorithmic}
\usepackage{graphicx}
\usepackage{textcomp}
\usepackage{xcolor}
\usepackage{soul}
\usepackage{comment}
\usepackage[linesnumbered,ruled,vlined]{algorithm2e}
\usepackage{balance}
\usepackage{microtype}
\usepackage{multirow}
\usepackage[normalem]{ulem}
\usepackage{soul}
\usepackage{lipsum}
\usepackage{wrapfig}
\usepackage{graphicx}

\definecolor{amethyst}{rgb}{0.6, 0.4, 0.8}

\newcommand{\toolname}{\textbf{CAREForMe}}

\begin{document}

\title{\toolname: Contextual Multi-Armed Bandit Recommendation Framework for Mental Health}

\author{Sheng Yu}
\email{centaurusyu2022@gmail.com}
\affiliation{%
  \institution{University of Southern California}
  \city{Los Angeles}
  \state{CA}
  \country{USA}
}

\author{Narjes Nourzad}
\email{nourzad@usc.edu}
\orcid{0009-0008-6315-8282}
\affiliation{%
  \institution{University of Southern California}
  \city{Los Angeles}
  \state{CA}
  \country{USA}
}

\author{Randye J. Semple}
\email{semple@med.usc.edu}
\orcid{0000-0001-6469-2032}
\affiliation{%
  \institution{University of Southern California}
  \city{Los Angeles}
  \state{CA}
  \country{USA}
}

\author{Yixue Zhao}
\email{yzhao@isi.edu}
\orcid{0000-0003-3046-6621}
\affiliation{%
  \institution{USC Information Sciences Institute}
  \city{Arlington}
  \state{VA}
  \country{USA}
}

\author{Emily Zhou}
\email{emilyzho@usc.edu}
\orcid{0009-0009-5004-6588}
\affiliation{%
  \institution{University of Southern California}
  \city{Los Angeles}
  \state{CA}
  \country{USA}
}

\author{Bhaskar Krishnamachari}
\email{bkrishna@usc.edu}
\orcid{0000-0002-9994-9931}
\affiliation{%
  \institution{University of Southern California}
  \city{Los Angeles}
  \state{CA}
  \country{USA}
}

\begin{abstract}
The COVID-19 pandemic has intensified the urgency for effective and accessible mental health interventions in people's daily lives. Mobile Health (mHealth) solutions, such as AI Chatbots and Mindfulness Apps, have gained traction as they expand beyond traditional clinical settings to support daily life. However, the effectiveness of current mHealth solutions is impeded by the lack of context-awareness, personalization, and modularity to foster their reusability. This paper introduces \toolname, a contextual multi-armed bandit (CMAB) recommendation framework for mental health. Designed with context-awareness, personalization, and modularity at its core, \toolname~ harnesses mobile sensing and integrates online learning algorithms with user clustering capability to deliver timely, personalized recommendations. With its modular design, \toolname~ serves as both a customizable recommendation framework to guide future research, and a collaborative platform to facilitate interdisciplinary contributions in mHealth research. We showcase \toolname's versatility through its implementation across various platforms (e.g., Discord, Telegram) and its customization to diverse recommendation features. 
\end{abstract}




\maketitle

\section{Introduction}
\label{sec:intro}

Mental health issues significantly increased during the COVID-19 pandemic and remain a global public health concern~\cite{cullen2020mental}. 
Mobile Health (mHealth) solutions, such as AI Chatbots and Mindfulness Apps, have gained traction due to the widespread use of mobile devices~\cite{Stawarz2019UseOS, mobileappusage2023}. By reducing the traditional reliance on human experts such as therapists, mHealth interventions can greatly enhance the accessibility, availability, and affordability of mental health care. They bridge the gap between clinics and homes, enabling continuous mental health monitoring and treatment in users' daily lives.

While mHealth solutions offer promising benefits, they currently face three major challenges. First, they often lack \emph{context-awareness} to recognize when the user is in need of help. This poses a burden on the user to initiate the intervention when help is needed, 
which is often when users are least motivated to do so. 
Second, these solutions fall short in \emph{personalization}. Many mHealth apps have a slow start to understand a user's history and preferences and rely heavily on user input. Users are either provided with the same intervention (e.g., mindfulness modules) or are tasked to describe their situation and navigate to the desired module by themselves each time they seek assistance.
Lastly, despite many mHealth solutions sharing common features, such as sensor data collection and recommendation algorithms, they are often developed in an \emph{ad-hoc} manner, causing duplicated efforts in this research area.

To address these challenges, we propose \underline{\toolname}, a \underline{\textbf{C}}ontex\-tual multi-\underline{\textbf{A}}rmed bandit \underline{\textbf{RE}}commendation \underline{\textbf{F}}ramew\underline{\textbf{or}}k for \underline{\textbf{Me}}ntal health. \toolname~ is a customizable framework designed with \emph{context-awareness}, \emph{personalization}, and \emph{modularization} in mind, aiming to understand the user's context and provide the right recommendations at the right time. 
First, \toolname~ employs mobile sensing to monitor the user's change of states, aiming to identify \emph{when} the recommendations should be delivered.
Second, \toolname~ integrates user clustering and contextual multi-armed bandit (CMAB) algorithms~\cite{pmlr-v9-lu10a} that continuously learn from user feedback to improve \emph{what} to recommend based on user preferences. We choose CMAB algorithms because they are more adaptive when learning from new data without the need to retrain from scratch compared to matrix factorization~\cite{Xu2021BanditMFMB, elena2021survey}. CMABs are also better at handling new users, i.e., the so-called ``cold-start'' problem, compared to traditional methods such as collaborative filtering~\cite{elena2021survey}. Most importantly, CMABs are \emph{context-sensitive}, using real-time user feedback to dynamically adjust the recommendations based on user's situation.
Finally, \toolname~ is a customizable framework with a modular design that decouples the underlying components, allowing different approaches to ``plug and play'' in the framework, without requiring the domain knowledge of the entire framework pipeline. For example, a mental health expert can focus on tailoring the recommendation content, and reuse existing recommendation algorithms (e.g.,  \toolname's built-in CMAB algorithm) to deliver her interventions. Similarly, a mobile app developer can focus on implementing the front-end user interface (e.g., a mobile app) to interact with the user and deliver the recommendations obtained from \toolname's CMAB back-end. 
Thus, \toolname~ stands not only as a customizable recommendation framework, but also as a collaborative living repository, fostering interdisciplinary contributions and facilitating research in the area of mHealth.

\begin{figure*}
    \centering
    \includegraphics[width=\textwidth]{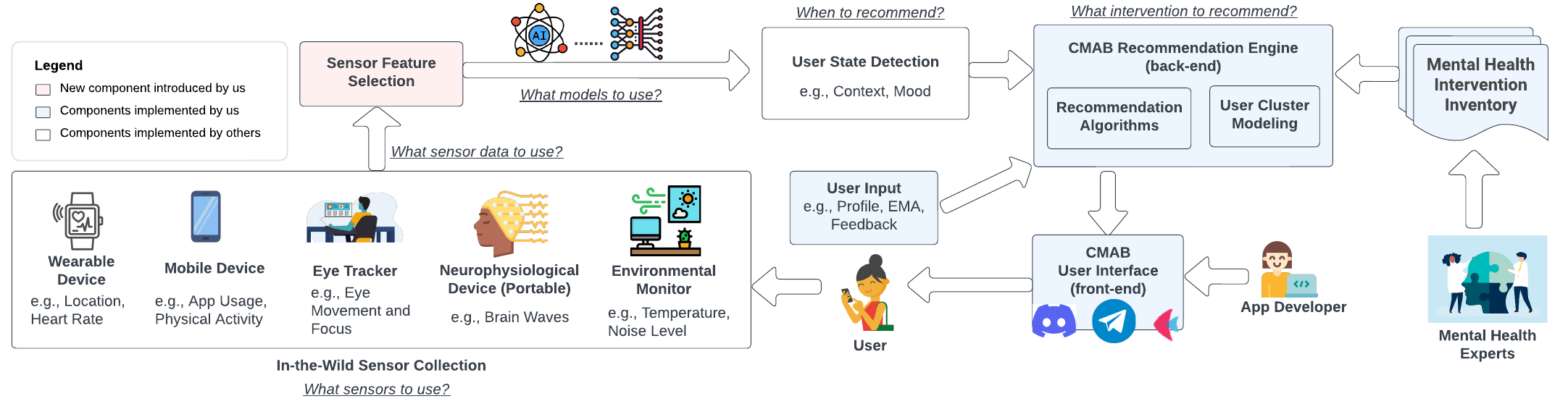}
    \caption{The overview of \toolname's framework design that is context-aware, personalized, and modular.}
    \label{fig:framework}
\end{figure*}

To demonstrate \toolname’s potential, we implement its core components with customizable features as a Chatbot recommendation system where the recommendations are mindfulness-based exercises and gratitude practices.
The recommendation content can be easily customized, reused, or extended, without requiring any technical knowledge.
Specifically, our recommendation system is implemented based on Upper Confidence Bound (UCB)~\cite{carpentier2011upper}, providing suggestions according to the user's context by continuously learning from the user's feedback.
Moreover, user clustering is implemented and can be optionally enabled in our system to provide suggestions based on the preferences of similar users instead of a particular user. This largely improves the recommendations for relatively new users with limited history to learn from, and also addresses the ``missing reward'' problem when the user forgets to provide the feedback~\cite{Bouneffouf2020ContextualBW}.
To show \toolname's flexibility and reusability, we implement 3 Chatbot systems that reuse the same back-end recommendation engine on Discord~\cite{discordbot}, Telegram~\cite{telegrambot}, and Flet~\cite{fletpwa} that functions on both Android and iOS systems.
A user simulation module is implemented to test out our Chatbot systems' functionalities including the user clustering capabilities.

We make the following contributions. 
(1) We design the first customizable CMAB-based framework of recommendation systems for mental health that is context-aware, personalized, and modular;
(2) We develop a flexible recommendation engine with user clustering mechanisms that allow different levels of customization;
(3) We implement an end-to-end recommendation system on 3 different Chatbot platforms, demonstrating the framework's flexibility and reusability;
(4) We make our artifacts publicly available~\cite{MABrepo}, directly improving the reproducibility and reusability of our research, and fostering future research in this area.

The rest of the paper is organized as follows.
Section~\ref{sec:design} introduces the design of \toolname~ framework and its modular components. Section~\ref{sec:implementation} describes our  implementation of \toolname's core components and details its customizable capabilities. Section~\ref{sec:related} discusses related work in this area, and Section~\ref{sec:discussion} elaborates our vision for adopting \toolname~ in practice and future work ahead.
\section{Framework Design}
\label{sec:design}

This section describes \toolname's design that incorporates \emph{context-awareness}, \emph{personalization}, and \emph{modularization} as discussed earlier.
Our design aims to provide a comprehensive, yet flexible framework that serves as a standard to guide the research in this area and improve the reusability of each other's work. 
Figure~\ref{fig:framework} shows the overview of \toolname~ framework with 7 components. 
These components are extracted from our extensive literature review in this area, and the core components in the CMAB recommendation pipeline are implemented by us (see Section~\ref{sec:implementation}). We now describe the high-level functionalities of each component.

\textbf{In-the-Wild Sensor Collection} guides the researchers to first answer \emph{what sensors to use} given their specific problem domain and resource constraints~\cite{wang2009framework}. We highlight ``in-the-wild'' as this paper focuses on providing recommendations during the user's daily life, as opposed to a lab setting. This is recently reported as an important but overlooked area~\cite{mello2023wild}. We identify 5 main categories of in-the-wild sensors: (1) \emph{Wearable Devices} can be worn in daily life, such as smartwatches and smart bands. They often monitor the user's location, heart rate, Heart Rate Variability (HRV), stress level, and activities such as walking; (2) \emph{Mobile Devices} refer to smartphones or tablets that can provide rich information about the user's states, including app usage and physical activities. For example, mobile sensors~\cite{AndroidSensors} can determine a user's activity, such as lying down, walking, or sitting. App usage data can offer insights into various aspects of a user's well-being, including sleep quality and nutritional content of meals; (3) \emph{Eye Tracker} devices are used to monitor and measure the position and movement of the eyes. They have a large potential to detect the user's mental state, such as emotion and fatigue~\cite{lim2020emotionEye,gunawardena2022eye}; (4) \emph{Neurophysiological Devices} can measure the user's neural activities in the brain to uncover insights into the user's mental states (e.g., cognitive abilities, stress level) via signals such as EEG~\cite{rayi2020eeg} and fNIR~\cite{pinti2020presentfNIRS}. As \toolname~ focuses on \emph{in-the-wild} sensors, we specifically target \emph{portable} devices that can be used outside of lab settings~\cite{larocco2020systemicEEG,artinis,Muse}; (5) \emph{Environmental Monitors} measure the user's surrounding environment (e.g., work, home) to provide the environmental context of the user, such as the surrounding temperature, noise level, and lighting. 

\textbf{Sensor Feature Selection} aims to answer \emph{what sensor data to use} as the input to the models that detect the user's state (e.g., mood) given the available sensors and the context. To the best of our knowledge, this is a novel component we introduce that is not considered by prior work. Existing work in this area relies on a fixed set of sensors' data as features to perform affect detection and has not explored context-informed sensor feature selection~\cite{bota2019review}.
\toolname~ incorporates \emph{context-awareness} in its Sensor Feature Selection to use suitable sensor data based on the user's context, aiming to increase the accuracy of the user state detection. For instance, when a user is sitting and working at a computer, the Eye Tracker's data about their focus and attention is likely to be more insightful than the data collected from their Mobile Device.

\textbf{User State Detection} takes the selected sensor features as input and answers \emph{what models to use} to detect the user's state, such as their mood. This information is used by the Recommendation Engine to determine \emph{when} certain intervention should be recommended to the user. For example, prior work focuses on detecting loneliness via supervised learning algorithms (e.g., logistic regression, gradient boosting classifier) using daily features such as communication, phone usage, and sleep behaviors collected from smartphones and smart bands~\cite{doryab2019loneliness}. If integrated into \toolname's User State Detection, other researchers can reuse such component to develop a Recommendation Engine that suggests social activities based on the user's loneliness level. Note that \toolname~ is a conceptual framework that does not assume any particular implementation of its components. Thus, the models used in the User State Detection can be either supervised or unsupervised with any architecture or algorithm of the researcher's choice.

\textbf{CMAB Recommendation Engine} is the core component in \toolname, integrating the recommendation algorithms and user clustering modeling to offer personalized suggestions. It takes 3 types of input: 
(1) the output from the User State Detection to determine \emph{when} to recommend; 
(2) the \textbf{User Input} such as user profile (e.g., user personality, age), ecological momentary assessment (EMA) of the user's behaviors and mental states, and user feedback of \toolname's suggestions which is used by the recommendation algorithms to \emph{dynamically} improve its results;
and (3) the \textbf{Mental Health Intervention Inventory}, often tailored by mental health experts in a specific domain, such as psychiatrists focusing on stress, anxiety, or loneliness.
The output of CMAB Recommendation Engine is used by the \textbf{CMAB User Interface} to deliver intervention suggestions to the user and obtain their feedback. This component is often implemented by app developers specialized in user interface design and development, without requiring the technical knowledge of the recommendation systems. For example, we have implemented 3 Chatbot CMAB User Interfaces on different platforms (e.g., Discord, Telegram) using the same CMAB Recommendation Engine to show \toolname's flexibility and reusability (see Section~\ref{sec:implementation}).
Our modular design adopts the ``separation of concerns'' principle~\cite{kulkarni2003separation}, enabling collaborative contributions from experts across various disciplines.
\section{Framework Instantiation}
\label{sec:implementation}

This section details our recommendation system that implements \toolname's core components, focusing on \emph{what} intervention to recommend.
As discussed earlier, the remaining components focus on \emph{when} to recommend by leveraging sensor data to detect the user state. We did not implement such components as they fall into a crowded research area and can be implemented by reusing existing work in various problem domains such as stress detection~\cite{cho2019instantStress, chikersal2021detectingDepression}.
We now detail the implementation of our recommendation system.

\textbf{Mental Health Intervention Inventory} is implemented in \texttt{YAML} format~\cite{yamlfile}, specifically designed for mental health experts such as psychiatrists, psychologists, or social workers without requiring any technical knowledge about recommendation systems. Mental health experts can configure the \texttt{YAML} file to specify the number of interventions to recommend each time, and the interventions with their corresponding context. Each intervention is defined as (\textit{title}, \textit{description}, \textit{image}, \textit{context}), such as (``\textit{STOP}'', ``\textit{Stop, Take a deep breath, Observe, and Proceed.}'', ``\textit{image.png}'', ``\textit{home}''). Our implementation uses mindfulness-based interventions and gratitude practices that are shown to be effective for people's daily lives~\cite{semple2019mindfulness}. 
Note that the inventory can grow to include interventions across various areas (e.g., anxiety, depression, anger management), and the system can be configured to use a subset of the interventions for targeted mental health issues.
The current interventions can be found in our public repository~\cite{MABrepo}, which are highly reusable and extensible.

\textbf{CMAB Recommendation Engine} is designed to continuously learn and refine its suggestions based on real-time user feedback from \textbf{User Input}. Its recommendation algorithm is implemented based on Upper Confidence Bound (UCB)~\cite{carpentier2011upper}, designed to provide context-sensitive recommendations, ensuring that the suggestions are tailored to individual user's context (e.g., home, work). 
The user clustering modeling is implemented using K-means~\cite{ahmed2020kmeans},
which can further refine the suggestions by learning from similar users' data. 
Note that there is a trade-off between the amount of data to learn from and the level of personalization achieved. Using data from all users maximizes the amount of training data but may reduce recommendation specificity, while using only individual user data ensures tailored recommendations but may require extended learning time.
Our system can be customized to achieve either of the scenarios above, or the middle ground by enabling the user clustering modeling to learn from a subset of the users that belong to the same cluster.
Furthermore, our modular design allows other clustering techniques to be easily ``plugged in'' by only replacing user clustering modeling without the need to change the rest of the system, such as replacing K-means with ADCB clustering~\cite{wang2023adcb}.


\begin{algorithm}[t!]
\begin{spacing}{0.85}
\small
\DontPrintSemicolon 
\KwIn{User $user$, Context $context$, Set $interventions$, \\ 
\qquad \quad Vector $CMAB\_score$, Vector $cluster$, Bool $implicit$}

\If{$user$.session\_num == 0}{
    $reco\_set \gets \textsc{SuggestAll}(interventions, CMAB\_score, context)$
}
\ElseIf{$user$.session\_num < $threshold$}{
    $reco\_set \gets \textsc{Suggest}(interventions, CMAB\_score, context)$
}
\Else{
    $user.cluster \gets \textsc{GetCluster}(user, cluster) $ \\
    $reco\_set \gets \textsc{Suggest}(interventions, CMAB\_score, context, user)$
}

$choice \gets \textsc{PromptUserForSelection}(reco\_set)$\\
$feedback \gets \textsc{PrompUserForFeedback}(choice)$\\

\If{$implicit$ == TRUE}{
    \textsc{UpdateCMAB}($user$, $context$, $choice$, $feedback$, $reco\_set$)\\
}
\Else{
    \textsc{UpdateCMAB}($user$, $context$, $choice$, $feedback$)\\
}

\textsc{UpdatePrefVector}($user$, $context$, $choice$, $feedback$)\\
\textsc{UpdateCluster}($user$, $context$, $choice$, $feedback$, $cluster$)\\

\caption{\sc CMAB Recommendation Process}
\label{alg:user_recommendation}
\end{spacing}
\end{algorithm}

Our system's recommendation process is implemented as a Chatbot to interact with the user, providing suggestions and obtaining corresponding user feedback.
Algorithm~\ref{alg:user_recommendation} details one session of the recommendation process with user clustering enabled. The input includes the information of the $user$ maintained by our system (e.g., how many sessions the user has interacted with the system), 
the $context$ the user is in (e.g., work), a set of $interventions$ to recommend from, the vector of the $CMAB\_score$ that stores the information of all of the user's past preferences that will be used to rank the recommendations~\cite{carpentier2011upper}, the vector $cluster$ that stores the information of similar users' past preferences in each cluster, and whether user's $implicit$ preferences on the unchosen suggestions should be considered when calculating the $CMAB\_score$.
After the session starts, the number of how many sessions the user has interacted with the system ($user.session\_num$) is checked. 
If the user is new, the suggestions ($reco\_set$) are learned from the preferences of \emph{all} the users in the system, and the $interventions$ with the highest $CMAB\_score$ in the user's provided context will be suggested (Line 2). This addresses the ``cold-start'' problem and ensures that new users are presented with popular suggestions across all the users. 
If $user.session\_num$ is smaller than a customizable threshold, the user gets the suggestions ($reco\_set$) from the $interventions$  based on their own past preferences, i.e., the $CMAB\_score$ calculated based on their past sessions using UCB method~\cite{carpentier2011upper} (Line 4). 
When $user.session\_num$ goes beyond the threshold, the $user$ is assigned to the closest cluster based on the Euclidean distance between the $cluster$ vector and the vector representing their past preferences (Lines 6-7). 
In this case, the $reco\_set$ are calculated by using the $CMAB\_score$ of similar users who belong to the same cluster.


Next, the recommendations are presented to the user to get the user input on which suggestion to choose ($choice$) and the $feedback$ for such suggestion using a score between 0 to 5 (Lines 8-9).
The obtained user input is used to update the $CMAB\_score$ for this particular user (Lines 10-13). If $implicit$ is set to true, not only the $CMAB\_score$ for the $choice$ will be updated, but the $CMAB\_score$ for the remaining suggestions in the $reco\_set$ will also be updated based on a customizable value. For instance, one may consider the unselected suggestions unfavored by the user and set a negative value to update the $CMAB\_score$ for those suggestions.
Finally, the user's preference vector that maintains the history of the user's preferences is updated accordingly, including their choices and feedback (Line 14).
The $cluster$ vector is updated as well to include the current session's data in the user's cluster (Line 15).


\textbf{CMAB User Interface} uses the output from  \textbf{CMAB Recommendation Engine} to deliver the recommendations to the users and obtain their feedback. It is designed to be decoupled from the underlying logic of the recommendation engine to enable the flexibility for app developers to customize their desired user interface. As a demonstration of such flexibility, we implement 3 variations of  \textbf{CMAB User Interface} using the same \textbf{CMAB Recommendation Engine}.
Specifically, 3 Chatbot systems are implemented on Discord~\cite{discordbot}, Telegram~\cite{telegrambot}, and mobile platforms using Flet framework~\cite{fletpwa} which can function on both Android and iOS platforms.



 




\section{Related Work}
\label{sec:related}

To the best of our knowledge, \toolname~ is the first framework that provides a comprehensive view throughout the recommendation pipeline based on CMAB that is context-aware, personalized, and modular. The closest work~\cite{gonul2018jitaiTemplate} proposed a template-based digital intervention design mechanism that supports the configuration of just-in-time adaptive interventions (JITAIs) for chronic health issues. It allows developers to instantiate its main components of decision points, intervention options, tailoring variables, and decision rules. However, this framework does not adapt to real-time user feedback and does not consider various types of sensors.


\textbf{Mobile and Wearable Sensing} for affect detection has become increasingly popular due to their ubiquity and unobtrusive nature that allow for a greater volume of data collection across multiple modalities~\cite{schmidt2019wearableReview, politou2017survey}, such as detecting loneliness using data from smartphones and smart bands~\cite{doryab2019loneliness}. 
However, these efforts were achieved separately, each requiring the development of its own software pipelines. This directly motivates us to design a customizable and modular framework to facilitate future work in this domain.

\textbf{Health Recommender Systems (HRS)} utilizes machine learning (ML) techniques and mobile platforms to support human health and wellness in many areas, such as providing recommendations for individuals with mental illness~\cite{rohani2020MUBS}.
In recent years, multi-armed bandits (MAB) algorithms have found success in applications such as weight loss interventions \cite{zhou2023MAB}, emotion regulation for mental illness \cite{ameko2020MAB}, and personalized physical activity recommendations \cite{rabbi2017}. However, HRSs often rely on an individual's health history, demographic information, and self-reports, and do not focus on delivering in-the-wild, real-time adaptive interventions~\cite{croos2021recSystemReview}. 


\textbf{JITAI Systems} as mentioned above have been used to provide real-time personalized support to promote positive behavior change for the improvement of health in many areas \cite{nahum2017jitaiReview}, such as mental health \cite{ben2013FOCUS, coppersmith2022jitaiSuicide} and alcohol usage \cite{gustafson20214alcoholism}. 
JITAIs aim to provide the right intervention based on the user's change of state and context in real-time through mobile sensing technologies \cite{nahum2017jitaiReview}, and have significantly outperformed non-JITAI treatments \cite{wang2020jitaiReview}. \toolname~ falls into the JITAI category for mental health, providing a comprehensive yet flexible framework to allow different levels of customization that goes beyond prior work.


\section{\toolname's Vision}
\label{sec:discussion}

Our long-term vision for \toolname~ is three-fold.
First, \toolname's design serves as a reference architecture, providing a comprehensive view of the state-of-the-art recommendation pipeline to guide newcomers in this field. Specifically, our framework guides new researchers to answer a set of research questions and decide which component(s) are of interest. 
For instance, those interested in \emph{when} to recommend can delve into the User State Detection component, while others may contribute to the novel Sensor Feature Selection component we identified, an area previously unexplored.
Second, our implementation of \toolname's core components showcases the framework's utility across various disciplines. It demonstrates how mental health professionals, app developers, and algorithm experts can leverage the modular design of \toolname~ to concentrate on their strengths and build upon each other's work, directly fostering an environment of interdisciplinary collaboration.
Finally, as others in the community adopt our framework, we envision \toolname~ growing into a living repository for just-in-time adaptive interventions (JITAIs), potentially extending its reach beyond mental health. For instance, a fitness expert can tailor the suggestions using physical exercises, and the User State Detection can focus on the user's physical activity levels. Similarly, other potential applications of \toolname~ include providing context-sensitive medication reminders or nutritional advice.

In the future, we plan to incorporate continuous learning throughout the framework, enabling CMAB techniques to answer not only \emph{what}, but also \emph{when} to recommend.
As users have different levels of acceptance on the frequency of recommendations, CMAB can be applied to improve the intervention's timing by learning from user feedback. 
Furthermore, we aim to evolve \toolname~ into a collaborative repository, similar to DBDP project~\cite{bent2021DBDP}, but extend to the whole recommendation system pipeline.
We plan to migrate existing work into \toolname, thereby enhancing their reusability and facilitating new discoveries. As the repository populates, the scope will grow to include capabilities such as fine-grained privacy control on the user's data collection, an access control model for data sharing, and a platform to connect researchers from various disciplines within the \toolname~ ecosystem. 
We believe the connection across disciplines is essential to advance research in human well-being given the intrinsic link between physical and mental health. \toolname~ has the potential to bridge the gap by providing a holistic view of user health with cross-disciplinary insights and adaptive recommendations that span different areas to improve the user's overall well-being, including mindfulness exercises, physical activities, nutritional advice, and so on.

\bibliographystyle{ACM-Reference-Format}
\bibliography{reference}

\end{document}